\documentclass[a4paper,conference]{IEEEtran}
\ifCLASSINFOpdf
\else
\fi
\usepackage{subfiles}
\usepackage{ amssymb }
\usepackage{multirow}
\usepackage{enumitem}
\usepackage{xcolor}
\usepackage{graphicx}
\usepackage{xspace}
\usepackage{float}
\makeatletter
\DeclareRobustCommand\onedot{\futurelet\@let@token\@onedot}
\def\@onedot{\ifx\@let@token.\else.\null\fi\xspace}

\def\etc{\emph{etc}\onedot}

\def\etal{\emph{et al}\onedot}

\makeatother
\DeclareMathAlphabet\mathbfcal{OMS}{cmsy}{b}{n}

\newcommand\blfootnote[1]{%
  \begingroup
  \renewcommand\thefootnote{}\footnote{#1}%
  \addtocounter{footnote}{-1}%
  \endgroup
}

\begin{document}
\title{Jointformer: Single-Frame Lifting Transformer with Error Prediction and Refinement for 3D Human Pose Estimation}

% \author{
% \IEEEauthorblockN{Sebastian Lutz$^\ast$}
% % \IEEEauthorblockA{V-SENSE\\
% % Trinity College Dublin\\
% % Dublin, Ireland\\
% % Email: lutzs@tcd.ie}
% \and
% \IEEEauthorblockN{Richard Blythman$^\ast$}
% % \IEEEauthorblockA{V-SENSE\\
% % Trinity College Dublin\\
% % Dublin, Ireland\\
% % Email: blythmar@tcd.ie}
% \and
% \IEEEauthorblockN{Koustav Ghosal}
% % \IEEEauthorblockA{V-SENSE\\
% % Trinity College Dublin\\
% % Dublin, Ireland\\
% % E-mail: ghosalk@tcd.ie}
% \and
% \IEEEauthorblockN{Matthew Moynihan}
% % \IEEEauthorblockA{V-SENSE\\
% % Trinity College Dublin\\
% % Dublin, Ireland\\
% % E-mail: mamoynih@tcd.ie}
% \and
% \IEEEauthorblockN{Ciaran Simms}
% % \IEEEauthorblockA{V-SENSE\\
% % Trinity College Dublin\\
% % Dublin, Ireland\\
% % E-mail: csimms@tcd.ie}
% \and
% \IEEEauthorblockN{Aljosa Smolic}
% % \IEEEauthorblockA{V-SENSE\\
% % Trinity College Dublin\\
% % Dublin, Ireland\\
% % E-mail: smolica@tcd.ie}}
% }
\author{
\IEEEauthorblockN{$^1$Sebastian Lutz$^\ast$, $^{1,2}$Richard Blythman$^\ast$, $^{1,3}$Koustav Ghosal, $^1$Matthew Moynihan, $^2$Ciaran Simms, $^1$Aljosa Smolic}
\IEEEauthorblockA{$^1$School of Computer Science and Statistics,
Trinity College Dublin\\
$^2$School of Mechanical, Manufacturing and Biomedical Engineering,
Trinity College Dublin\\
$^3$Accenture Labs\\
% Dublin, Ireland\\
\{lutzs, blythmar, ghosalk, mamoynih, csimms, smolica\}@tcd.ie}
}


%Telephone: (800) 555--1212, Fax: (888) 555--1212}


\maketitle

\begin{abstract}

Monocular 3D human pose estimation technologies have the potential to greatly increase the availability of human movement data. The best-performing models for single-image 2D-3D lifting use graph convolutional networks (GCNs) that typically require some manual input to define the relationships between different body joints. We propose a novel transformer-based approach that uses the more generalised self-attention mechanism to learn these relationships within a sequence of tokens representing joints. 
% Our architecture consists of several transformer encoders in which each joint of the pose is treated as a token. 
% This allows our network to share position information intelligently between joints and to propagate salient information to subsequent transformers. 
We find that the use of intermediate supervision, as well as residual connections between the stacked encoders benefits performance. 
We also suggest that using error prediction as part of a multi-task learning framework improves performance by allowing the network to compensate for its confidence level. 
We perform extensive ablation studies to show that each of our contributions increases performance.
Furthermore, we show that our approach outperforms the recent state of the art for single-frame 3D human pose estimation by a large margin. Our code and trained models are made publicly available on Github.

\end{abstract}
\blfootnote{$^\ast$ denotes equal contribution}
\section{Introduction}
\label{sec:intro}

Movement analysis is vitally important for applications including mixed reality, human-computer interaction, sports biomechanics and physiotherapy. 
% maximising sporting performance in athletes and for ensuring a high quality of life in the general population. 
% The general role of physiotherapists and coaches is to teach high-quality movement to reduce injury risk. 
However, human motion is highly complex and subjective, and the high dimensionality and the variation between subjects means that much is still not understood. While motion capture systems have been used to quantify movement since the 1980s \cite{davis1991gait}, the equipment is expensive, the datasets are largely constrained to laboratory settings and relatively few are publicly-available. At the same time, the internet has collected vast amounts of \textit{in-the-wild} (unlabelled and unconstrained) images and videos of moving humans. The maturation of monocular 3D human pose estimation (HPE) technologies has the potential to create a step-increase in data available and scenarios that can be assessed, which can ultimately be used to improve our understanding of human movement. 

% give a measure of confidence of 3D pose estimates in real applications. 

Monocular 3D human pose estimation involves the prediction of 3D joint positions from a single viewpoint. While video techniques can leverage temporal information to improve accuracy, single-image estimators remain useful in their own right. For example, useful information about human movement can be learned from online image datasets or video datasets with low frame-rate. Furthermore, some video approaches opt to use a combination of single-frame spatial models along with a multi-frame temporal model \cite{zheng20213d}. Thus, strong single-image 3D pose lifters can also improve the performance on video data.

Direct estimation techniques aim to estimate 3D human pose directly from images \cite{pavlakos2017coarse}. However, diverse image datasets with 3D pose labels are sparse, and it is convenient to leverage the high accuracy of off-the-shelf 2D pose estimators \cite{cao2019openpose, sun2019deep} that are well-suited for the pixel-processing task. Lifting approaches predict the 3D pose from 2D joint predictions provided from such estimators. The types of neural network architectures used to learn this mapping have progressively evolved. The original simple baseline \cite{martinez2017simple} for pose lifting used a multi-layer perceptron to achieve surprising accuracy, even without information from image features. More recent works have highlighted that treating pose as a vector ignores the spatial relationships between joints, and that graph structures may be better-suited \cite{zhao2019semantic}. However, existing works on graph convolutions require a hand-crafted adjacency matrix to define the relationship between nodes. Also, standard graph convolutions model the relationships between neighboring joints with a shared weight matrix, which is sub-optimal for modelling articulated bodies since the relations between different body joints may be different \cite{DBLP:conf/eccv/LiuDZ0T20}. Furthermore, stacking GCN layers may result in over-smoothing \cite{zhou2020towards}. In contrast, the self-attention operator of the transformer model generalises the feed-forward layer to be dynamic to the input, and the relationship between joints can be learned rather than manually encoded.
% Compared with graph convolutions, the relationship between joints can be learned by the self-attention mechanism.

Transformers first unseated recurrent neural networks as the predominant models in natural language processing (NLP) \cite{vaswani2017attention}, and have recently gained success in replacing convolutional neural networks in vision tasks such as image classification \cite{dosovitskiy2020image}, object detection \cite{carion2020end}, and action recognition \cite{sharir2021image}. However, very few studies have applied transformers to the task of 3D human pose estimation to date. Existing works have either used a direct estimation approach \cite{lin2020end} or focused their studies on video-based 2D-3D lifting \cite{zheng20213d, li2021lifting}. 
% the considerations such as ... and memory requirements are unique. 
To our knowledge, ours is the first method to adopt a transformer for single-image 2D-3D pose lifting. The input and output tokens of the sequence represent joints, and thus we refer to our method as the Jointformer.

% Intermediate Supervision
% Works that use intermediate supervision. ... Human pose estimation. stacked hourglass \cite{newell2016stacked}. Finally, DETR \cite{carion2020end} used intermediate supervision between the stacked decoder blocks. 

% Error Prediction
% It helps with training the same way multi-task training does. And it gives additional output that might be useful. if you know that the prediction is very uncertain for this one limb. It could be useful depending on what you do with the prediction

Accordingly, our contributions are as follows:
\begin{itemize}[topsep=1pt,itemsep=-1ex,partopsep=1ex,parsep=1ex]
    \item We present a novel single-frame 2D-3D lifting Joint Transformer for human pose estimation. 
    \item We suggest that predicting the error associated with each of the joints improves accuracy, by enabling the network to compensate for its own uncertainty.
    \item We propose a Refinement Transformer to refine the 3D pose predictions of the Joint Transformer, based on the 3D prediction themselves, the input 2D joints, and the predicted errors for each joint.
    \item We introduce intermediate supervision on the 3D joint and error predictions within the stack of transformer encoders, facilitated by linear layers. 
    \item We show that this architecture achieves state-of-the-art results on the Human3.6M and MPI-INF-3DHP datasets for single-frame 3D human pose estimation.
\end{itemize}
\section{Related Works}
\label{sec:related_works}

% \subsection{Human Pose Estimation}

\textbf{Human Pose Estimation:} Due to its applicability in diverse areas such as action recognition, augmented and mixed reality \etc, HPE has emerged as a very active problem in computer vision.
% and the number of publications abound. 
We present the literature relevant to
% five different categories ---
direct estimation and 2D-3D lifting methods.
% (1) Direct estimation from images  (2) Estimation from 2D keypoints (3) Monocular or single-view (4) Multi-view (5) Temporal. Note, that these categories are not mutually exclusive and most methods are usually a combination of the aforementioned categories.
Direct estimation refers to estimating the 3D pose directly from raw images. Pavlakos \etal \cite{DBLP:conf/cvpr/PavlakosZD18} use the ordinal depths of human joints to provide a weaker supervision signal.
% Pavlakos \etal  \cite{pavlakos2017coarse} present an end-to-end and coarse-to-fine approach for training a CNN that predicts per voxel likelihoods for each joint. Sun \etal \cite{Sun_2018_ECCV} introduce a differentiable integral operator for heat-map based 3D pose estimation. 
Sun \etal \cite{DBLP:conf/iccv/0001SLW17} exploit the joint connection structure and define a compositional loss function that encodes long range interactions of the pose. 
Martinez \etal \cite{martinez2017simple} explored decoupling of the problem into 2D human pose estimation and 2D-3D lifting and used a vanilla neural network to learn the mapping. Zhao \etal \cite{zhao2019semantic} exploit the spatial relations of nodes (both local and global) in a semantic graph convolutional network (SemGCN) to improve performance.
Ci \etal \cite{ci2019optimizing} introduced locally-connected networks to overcome the limited representation power of GCNs for estimating 3D pose.
Xu \etal \cite{DBLP:journals/corr/abs-2103-16385} use a graph stacked hourglass network to process graph-structured features across three different scales of human skeletal representations. 
Tekin \etal \cite{DBLP:conf/iccv/TekinMSF17} design a network to combine the strengths and weaknesses of direct estimation and lifting. Yang \etal \cite{yang20183d} propose a method for in-the-wild HPE using adversarial learning aided by a geometry-aware discriminator. 
% They model the pairwise relative locations and distances between body joints using a geometric descriptor which is used to boost the discriminator.
Fang \etal \cite{DBLP:conf/aaai/FangXWLZ18} present a lifting framework based on bidirectional recurrent networks which explicitly models auxiliary information such as kinematics, symmetry and motor coordination. Sharma \etal \cite{DBLP:conf/iccv/SharmaVBSJ19} use variational autoencoder for predicting a set of candidate 3D poses and ranks them using an ordinal score or oracle. Liu \etal \cite{DBLP:conf/eccv/LiuDZ0T20} - in a comprehensive review of HPE - study different weight sharing strategies for graph convolution based lifting methods. Zhou \etal \cite{DBLP:conf/iccv/ZhouHJJL19} propose a two-stage method in which the 2D points are mapped to an intermediate latent space, followed by a volumetric regression to the 3D space.

\textbf{Transformers for Vision:} Convolutional neural networks (CNNs) \cite{lecun1989backpropagation} have remained the dominant model in computer vision for three decades. Since the convolution operates within small local neighbourhoods, deeply stacking convolutions becomes necessary to form large receptive fields and capture long-distance dependencies. Inspired by the self-attention mechanism in NLP \cite{vaswani2017attention}, the non-local neural network \cite{wang2018non} used a small number of non-local blocks between various stages of a CNN to better capture long-distance relationships for action recognition. Ramachandran \etal \cite{ramachandran2019stand} explored using stand-alone self-attention within small windows as a basic building block (rather than an augmentation on top of convolutions) for image classification and object detection. Similarly, Zhao \etal \cite{zhao2020exploring} used a self-attention network with pairwise and patchwise self-attention for image recognition. 

Rather than basing the structure of building blocks on ResNet-style architectures, the vision transformer (ViT) \cite{dosovitskiy2020image} directly applied a standard transformer (with as few modifications as possible) to classify images. Images were input as sequences of flattened patches, with patches treated similar to tokens (words) in NLP applications. Similar approaches have since been applied to video classification \cite{arnab2021vivit} and action recognition \cite{sharir2021image}.

Despite showing significant promise in other vision tasks, few studies have applied transformers to the task of 3D human pose estimation. Liu \etal were the first to use an attention mechanism to adaptively identify significant frames in temporal windows of human motion videos. Lin \etal \cite{lin2020end} used a transformer for direct estimation of 3D human pose and shape for a single input frame. The input tokens consisted of a feature vector extracted with a CNN that is concatenated with either randomly-initialised 3D joints or mesh vertices. The input input token dimensions were progressively reduced after each encoder. Zheng \etal \cite{zheng20213d} addressed the task of video-based 3D human pose estimation and incorporated a spatial transformer as a component within a larger temporal transformer, where the output tokens are encoded features rather than 3D joint positions. Thus, it cannot be applied to single-frame 2D-3D lifting and its performance was not evaluated on images. Li \etal \cite{li2021lifting} used a lifting transformer that replaced feed-forward layers in the encoder with strided temporal convolutions. It operated on a temporal sequence, where each input token was a whole skeleton. This means that useful information cannot be passed between joints in the skeleton.

% \subsection{Error Prediction, Refinement and Intermediate Supervision}

\textbf{Error Prediction, Refinement and Intermediate Supervision:}  The predictions from human pose estimation models inevitably contain errors. Ronchi \etal \cite{ruggero2017benchmarking} performed a rigorous analysis of the prediction errors of 2D HPE estimators on the COCO dataset \cite{lin2014microsoft}. The relative frequency and impact to performance of different categories of errors (jitter, inversion, swap, and miss) was investigated, and it was found that the errors showed similar distributions independent of the type of 2D pose estimator used. Moon \etal \cite{moon2019posefix} used these error statistics to generate synthetic poses, which were input to a refinement network. Similarly, Chang \etal \cite{chang2019poselifter} used specific error distributions to synthesise 2D poses to be used for 2D-3D pose lifting. Fieraru \etal \cite{fieraru2018learning} introduced a pose refinement network that takes as input both the image and a given pose estimate. Zhang \etal \cite{zhang2019human} constructed a pose graph to consider the relationship between different keypoints during refinement. Wang \etal \cite{wang2020graph} used a graph pose refinement (GPR) module to refine the visual features based on the relationship between keypoints.

Intermediate supervision is the practice of using additional loss terms at various stages of deep neural networks to improve performance. In 2D human pose estimation, Newell \etal \cite{newell2016stacked} used repeated bottom-up/top-down processing with intermediate supervision with their stacked hourglass network. DETR \cite{carion2020end} used intermediate supervision between the stacked transformer decoder blocks for object detection.
% \vspace{-0.2cm}
\section{Method}
\label{sec:method}
In this section, we describe our proposed network architecture for 3D human pose estimation. Following previous 2D-3D pose lifting approaches \cite{martinez2017simple}, we use an off-the-shelf model to generate 2D poses from images, and use these predictions to estimate the corresponding 3D poses. The 3D poses are estimated in camera coordinates and centered on the pelvis joint. While most of the previous state of the art use multilayer perceptrons (MLPs) \cite{martinez2017simple} or GCNs \cite{zhao2019semantic}, we use transformers to predict and refine the 3D pose in two parts. First, our Joint Transformer estimates the 3D pose and the prediction error from the 2D pose of a single frame. Afterwards, our Refinement Transformer further improves this prediction by using the intermediate prediction and prediction error. This novel combination of networks allows us achieve state-of-the-art performance in single frame 3D pose estimation.

% Simple fully-connected layers \cite{martinez2017simple}
% Graph convs \cite{zhao2019semantic}
% Transfomers...
% Estimate in camera coordinates
% lack of 3d ground truth posture data for images in the wild
% following previous works, root-relative estimations, that is, predicting the joint coordinates w.r.t the pelvis joint

\begin{figure*}[!t]
    \centering
    \includegraphics[width=0.8\textwidth]{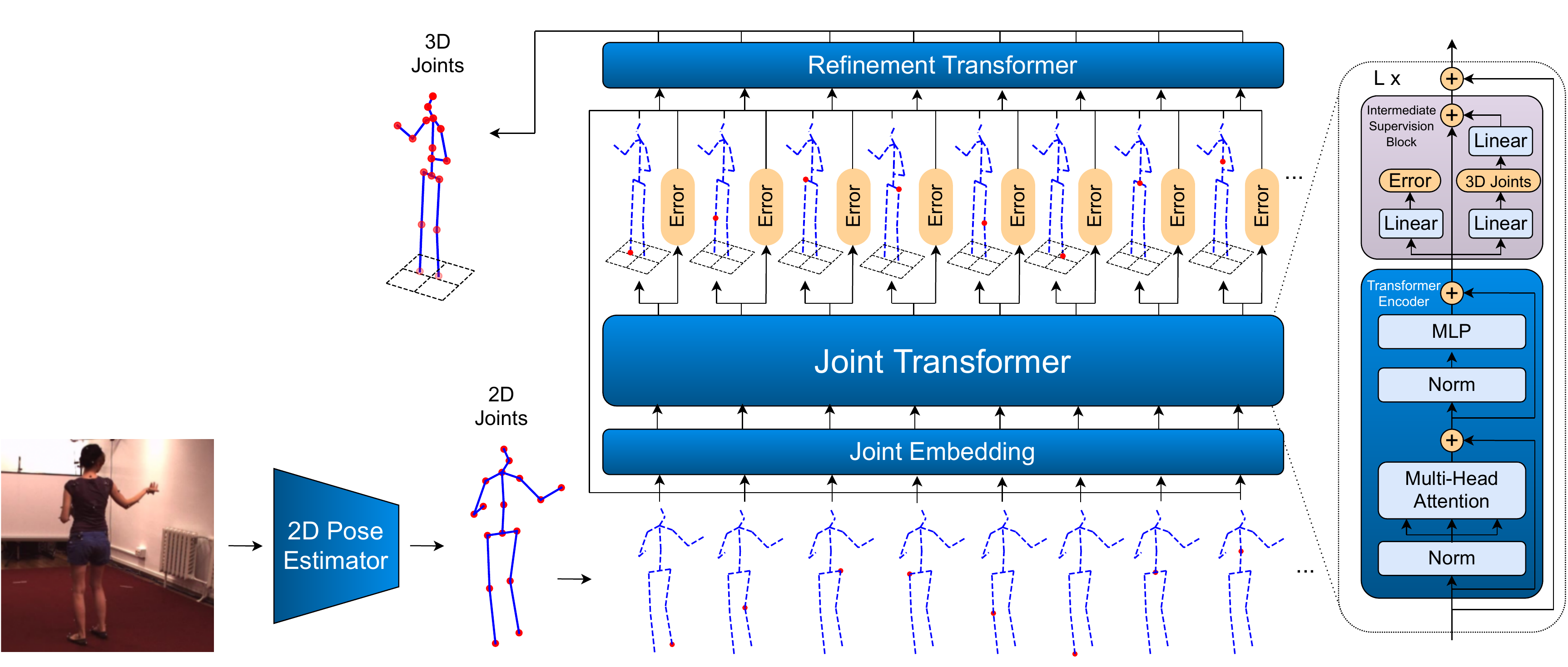}
    \caption{Our proposed Jointformer for single-frame 2D-3D human pose lifting.}
    \label{fig:jointtransformer}
\end{figure*}

\subsection{Joint Transformer}
\label{sec:joint_transformer}
Our Joint Transformer is designed to lift the input 2D pose to 3D for a single frame. Given a set of $J$ 2D joint coordinates $\mathbf{x}$, we consider each joint a token and first embed each into a higher dimension of size $c$: $f_{embed}(\mathbf{x}): \mathbb{R}^{J \times 2} \rightarrow \mathbb{R}^{J \times c}$. Traditionally in transformers, the embedding is followed by a positional encoding. In our case however, the order of joints is not changed during both training and testing, implicitly encoding the position in the input itself (\textit{i.e.} the first joint in the input is always the hip joint). This makes the addition of an explicit positional encoding redundant, and in fact hurts our prediction performance as we show in Section \ref{sec:ablation}. Following the embedding, we feed the joint tokens into a stack of transformer encoders where self-attention is applied to share information across each joint. We use the original implementation for the transformer encoder \cite{vaswani2017attention} and do not modify the hidden dimension further. Given $L$ transformer encoders, the output of each encoder is $\mathbf{z}_L \in \mathbb{R}^{J \times c}$. Finally, we regress the output of the last encoder $\mathbf{z}$ to predict the 3D pose using an MLP block consisting of layer normalization, dropout and a single linear layer $f_{pred}(\mathbf{z}): \mathbb{R}^{J \times c} \rightarrow \mathbb{R}^{J \times 3}$.

% The layers of a transformer show improved generalisation compared to architectures from previous works. Encoder vs decoder. The self-attention operator generalises the feed-forward layer to be dynamic on the input. This is achieved by using several fully-connected layers that operate. called key, value, query. Compared with graph convolutions, the relationship between joints can be learned by the self-attention mechanism. increased flexibility. 

\subsubsection{Intermediate supervision}
\label{sec:intermediate_supervision}
By design, the transformer encoder layers share information between tokens and learns which tokens are important for prediction. This gives even the first transformer encoder in our stack the ability (though not necessarily the capacity) to learn how joint tokens interact and to predict the 3D pose. We leverage this by training our network using intermediate supervision. Compared to previous methods, we do not compute the loss at the end of the network alone, but introduce a loss term after each individual transformer encoder in the stack. This allows the network to learn initial estimates that get further refined by each transformer encoder in the stack. Subsequent transformer encoders get passed down highly discriminative features from previous encoders and can focus on increasingly fine-grained features to refine previous estimates. To further help with this, we add residual connections around every encoder. In our implementation, each encoder is followed by an MLP predicting the 3D pose, to which a loss can be applied. We add an additional linear layer to the predicted 3D pose to embed the prediction back to the hidden dimension of the transformer and add the embedding to the original transformer output, as can be seen in the upper right of Figure \ref{fig:jointtransformer}.

\subsubsection{Pose embedding}
\label{sec:embedding}
The 2D input pose coordinates need to be expanded to the hidden dimension of our network. Previous methods used linear layers or graph convolutions for this task. In our method however, we use 1D convolutions with a kernel size of $1$ across all joints. This is equivalent to a linear layer that expands the 2D coordinates to the hidden dimension individually for each joint with weight sharing between each joint. Since the embedding weights are shared for each joint, the layer can focus on finding the best way to expand the 2D coordinates regardless of the location of the joint. Traditionally, transformers also require a positional encoding of the embedded tokens. Since the input order of joints never changes, the position is implicitly already encoded and additional encodings are redundant. We show in Table \ref{tab:ablation} that our method works better without any explicit positional encoding and our convolutional embedding outperforms a linear layer.

\subsubsection{Error prediction}
\label{sec:error_prediction}

Giving an estimate of the confidence in its prediction gives the network the ability to compensate. Since classification heads generate one prediction per class before a softmax layer, their prediction confidence can be directly inferred. In contrast, regression heads lack this ability out of the box.
% making it impossible to tell how confident the network is in its prediction.
% To facilitate error prediction, 
We add a second MLP mirroring the 3D pose prediction to the output of the transformer encoder, and force the network to implicitly learn its own uncertainty. Therefore, the error prediction will predict its own error per joint and per coordinate $f_{error}(\mathbf{z}): \mathbb{R}^{J \times c} \rightarrow \mathbb{R}^{J \times 3}$.
We observe that the addition of the error prediction stabilizes the training and leads to better overall results (Section \ref{sec:ablation}). It can also serve as important additional information when visualizing the pose or for practical application of the technology.

\subsubsection{Loss function}
\label{sec:loss}
The loss function which we use to train our network consists of two terms: the 3D prediction loss  $\mathcal{L}_{3D}$ and the error loss $\mathcal{L}_{error}$. For the 3D prediction loss, we use the mean-squared error between the predicted 3D poses $\mathbf{y}$ and the ground-truth poses $\hat{\mathbf{y}}$.
For the error loss, we first define the true error $\hat{\mathbf{e}}$ as the absolute difference between the predicted and the ground-truth pose $\hat{\mathbf{e}} = |\mathbf{y} - \hat{\mathbf{y}}|$. This allows us to use the mean-squared error between the predicted error $\mathbf{e}$ and the true error $\hat{\mathbf{e}}$ as the error loss.
Due to our intermediate supervision, we apply both of these loss terms after every transformer encoder in the stack. This leads to the combined loss:
\begin{equation}
    \mathcal{L}_{combined} = \frac{1}{L} \sum_{i=1}^L \frac{\mathcal{L}_{3D}^i + \mathcal{L}_{error}^i}{2},
\end{equation}
where $L$ is the number of transformer encoders in the stack.

% \subsection{Sequence prediction}
% \label{sec:sequence_prediction}
% How do we use our method for sequence prediction

\subsection{Refinement Transformer}
\label{sec:refinement}
In order to make further use of the error prediction and boost the accuracy of the pose prediction, we design an additional Refinement Transformer. This network is similar to our Joint Transformer with the following difference: the input is the original 2D pose concatenated to the predicted 3D pose and error prediction. We only stack two transformer encoders with a hidden dimension of $256$, and we do not use intermediate supervision. As with the Joint Transformer, the Refinement Transformer tokenizes each joint and uses a 1D convolution to embed the $8$-channel input to the hidden dimension. Following the stacked transformer encoders, we use a linear layer to predict the final 3D pose.

\subsection{Implementation details}
\label{sec:implementation_details}
We implement our approach in PyTorch \cite{DBLP:conf/nips/PaszkeGMLBCKLGA19} and use a single NVIDIA GeForce RTX 2080 TI for training. We train the Joint Transformer for $30$ epochs with a batch size of $256$ using the AdamW \cite{DBLP:conf/iclr/LoshchilovH19} optimizer, an initial learning rate of $0.001$ and cosine annealing learning rate decay \cite{DBLP:conf/iclr/LoshchilovH17}. All of our prediction layers also include dropout of $0.2$ during training. Afterwards, we fix the weights of the Joint Transformer and train the Refinement Transformer using the 3D prediction loss and employing the same hyperparameters. During evaluation we use test-time data augmentation by horizontal flipping, following \cite{pavllo20193d}.
\section{Experiments}
\label{sec:experiments}

In this section, we first introduce the datasets used to evaluate our proposed approach. The results are then compared to the state of the art, and an ablation study is performed.

\subsection{Datasets and Evaluation}

We consider two popular benchmarks for 3D HPE following recent works \cite{DBLP:journals/corr/abs-2103-16385}.

\textbf{Human3.6M} \cite{DBLP:journals/pami/IonescuPOS14}: This is the most widely-used dataset for 3D human pose estimation. It uses a motion capture system and 4 video cameras to capture the 3D pose information and accompanying images, respectively. The camera calibration parameters are used to project the 3D joint positions to the 2D image plane of each camera. The dataset contains 3.6 million images of 7 professional actors performing 15 generic movements such as walking, eating, sitting, making a phone call and engaging in a discussion. We follow the standard protocol, using subjects 1, 5, 6, 7 and 8 for training, and subjects 9 and 11 for evaluation.

\textbf{MPI-INF-3DHP} \cite{8374605}: This dataset contains images of three different scenarios: (i) studio with a green screen (GS), (ii) studio without green screen (noGS) and (iii) outdoor scene (Outdoor). We follow the common skeleton representation for both of these datasets, which contains $17$ joints.

Following common practice, for Human3.6M we evaluate by using the mean per-joint pixel error (MPJPE) in $mm$ between the ground truth and our predictions across all joints and cameras, after alignment of the root joint (\textit{Protocol \#1}). It is also common to align the prediction with the ground truth using a rigid transformation \cite{moreno20173d} (\textit{Protocol \#2}). For the MPI-INF-3DHP test set we use 3D-PCK (Percentage of Correct Keypoints) and AUC (Area Under the Curve) as evaluation metrics following previous works \cite{DBLP:conf/bmvc/LuoCY18, DBLP:journals/corr/abs-2103-16385, ci2019optimizing}.

\subsection{Comparison with State-of-the-Art Methods}
\begin{table*}[t]
\begin{center} 
\def\arraystretch{1.2} 
\resizebox{0.8\textwidth}{!}{

\begin{tabular}{ccccccccccccccccc}
\hline 
\textbf{(a) CPN keypoints} & \textbf{Direct} & \textbf{Discuss} & \textbf{Eating} & \textbf{Greet} & \textbf{Phone} & \textbf{Photo} & \textbf{Pose} & \textbf{Purch.} & \textbf{Sitting} & \textbf{SittingD.} & \textbf{Smoke} & \textbf{Wait} & \textbf{WalkD.} & \textbf{Walk} & \textbf{WalkT.} & \textbf{Avg.}\tabularnewline
\hline 
\hline 
Martinez \etal \cite{martinez2017simple} ICCV'17 & 51.8 & 56.2 & 58.1 & 59.0 & 69.5 & 78.4 & 55.2 & 58.1 & 74.0 & 94.6 & 62.3 & 59.1 & 65.1 & 49.5 & 52.4 & 62.9\tabularnewline
Tekin \etal \cite{DBLP:conf/iccv/TekinMSF17}  ICCV'17 &  54.2 & 61.4 & 60.2 & 61.2 & 79.4 & 78.3 & 63.1 & 81.6 & 70.1 & 107.3 & 69.3 & 70.3 & 74.3 & 51.8 & 63.2 & 69.7 \tabularnewline
Sun \etal \cite{DBLP:conf/iccv/0001SLW17} ICCV'17 (+) & 52.8 & 54.8 & 54.2 & 54.3 & 61.8 & 67.2 & 53.1 & 53.6 & 71.7 & 86.7 & 61.5 & 53.4 & 61.6 & 47.1 & 53.4 & 59.1\tabularnewline
Yang \etal \cite{yang20183d} CVPR'18 (+) & 51.5 & 58.9 & 50.4 & 57.0 & 62.1 & 65.4 & 49.8 & 52.7 & 69.2 & 85.2 & 57.4 & 58.4 & \textbf{43.6} & 60.1 & 47.7 & 58.6\tabularnewline
Fang \etal \cite{DBLP:conf/aaai/FangXWLZ18} AAAI'18 &  50.1 & 54.3 & 57.0 & 57.1 & 66.6 & 73.3 & 53.4 & 55.7 & 72.8 & 88.6 & 60.3 & 57.7 & 62.7 & 47.5 & 50.6 & 60.4\tabularnewline
Pavlakos \etal \cite{DBLP:conf/cvpr/PavlakosZD18} CVPR'18 (+) & 48.5 & 54.4 & 54.5 & 52.0 & 59.4 & 65.3 & 49.9 & 52.9 & 65.8 & 71.1 & 56.6 & 52.9 & 60.9 & 44.7 & 47.8 & 56.2\tabularnewline
Zhao \etal \cite{zhao2019semantic} CVPR'19 &  48.2 & 60.8 & 51.8 & 64.0 & 64.6 & \textbf{53.6} & 51.1 & 67.4 & 88.7 & \textbf{57.7} & 73.2 & 65.6 & 48.9 & 64.8 & 51.9 & 60.8\tabularnewline
Sharma \etal \cite{DBLP:conf/iccv/SharmaVBSJ19} ICCV'19 &  48.6 & 54.5 & 54.2 & 55.7 & 62.2 & 72.0 & 50.5 & 54.3 & 70.0 & 78.3 & 58.1 & 55.4 & 61.4 & 45.2 & 49.7 & 58.0\tabularnewline
Ci \etal \cite{ci2019optimizing} ICCV'19 (+) & 46.8 & 52.3 & \textbf{44.7} & 50.4 & 52.9 & 68.9 & 49.6 & 46.4 & 60.2 & 78.9 & 51.2 & 50.0 & 54.8 & 40.4 & 43.3 & 52.7\tabularnewline
Liu \etal \cite{DBLP:conf/eccv/LiuDZ0T20} ECCV'20 &  46.3 & 52.2 & 47.3 & 50.7 & 55.5 & 67.1 & 49.2 & \textbf{46.0} & 60.4 & 71.1 & 51.5 & 50.1 & 54.5 & 40.3 & 43.7 & 52.4\tabularnewline
Lin \etal \cite{lin2020end} CVPR'21 & - & -& -  & -& -  & -& -  & -& -  & -& -  & -& -  & -& - & 54.0\tabularnewline 
Xu \etal \cite{DBLP:journals/corr/abs-2103-16385} CVPR'21 & 45.2 & 49.9 & 47.5 & 50.9 & 54.9 & 66.1 & 48.5 & 46.3 & 59.7 & 71.5 & 51.4 & 48.6 & 53.9 & 39.9 & 44.1 & 51.9\tabularnewline
\hline
Base Transformer & 50.3 & 53.7 & 49.7 & 53.0 & 56.7 & 62.8 & 52.2 & 49.0 & 61.8 & 69.0 & 54.5 & 51.0 & 56.1 & 42.1 & 44.7 & 53.8 \tabularnewline
Ours & \textbf{45.0} & \textbf{48.8} & 46.6 & 49.4 & 53.2 & 60.1 & \textbf{47.0} & 46.7 & 59.6 & 67.1 & 51.2 & \textbf{47.1} & 53.8 & 39.4 & 42.4 & 50.5\tabularnewline
Ours + Refinement & 45.6 & 49.7 & 46.0 & \textbf{49.3} & \textbf{52.2} & 58.8 & 47.5 & 46.1 & \textbf{58.2} & 66.1 & \textbf{50.7} & 47.5 & 52.6 & \textbf{39.2} & \textbf{41.6} & \textbf{50.1}\tabularnewline

\hline 
\end{tabular}
}

\bigskip

\resizebox{0.8\textwidth}{!}{
\begin{tabular}{ccccccccccccccccc}
\hline 
\textbf{(b) GT keypoints} & \textbf{Direct} & \textbf{Discuss} & \textbf{Eating} & \textbf{Greet} & \textbf{Phone} & \textbf{Photo} & \textbf{Pose} & \textbf{Purch.} & \textbf{Sitting} & \textbf{SittingD.} & \textbf{Smoke} & \textbf{Wait} & \textbf{WalkD.} & \textbf{Walk} & \textbf{WalkT.} & \textbf{Avg.}\tabularnewline
\hline 
\hline 
Martinez \etal \cite{martinez2017simple} ICCV'17 &  45.2 & 46.7 & 43.3 & 45.6 & 48.1 & 55.1 & 44.6 & 44.3 & 57.3 & 65.8 & 47.1 & 44.0 & 49.0 & 32.8 & 33.9 & 46.8\tabularnewline
Zhou \etal \cite{DBLP:conf/iccv/ZhouHJJL19} ICCV'19 (+) & 34.4 & 42.4 & 36.6 & 42.1 & 38.2 & 39.8 & 34.7 & 40.2 & 45.6 & 60.8 & 39.0 & 42.6 & 42.0 & 29.8 & 31.7 & 39.9\tabularnewline
Ci \etal \cite{ci2019optimizing} ICCV'19 (+) & 36.3 & 38.8 & 29.7 & 37.8 & 34.6 & 42.5 & 39.8 & 32.5 & 36.2 & \textbf{39.5} & 34.4 & 38.4 & 38.2 & 31.3 & 34.2 & 36.3\tabularnewline
Zhao \etal \cite{zhao2019semantic} CVPR'19 &  37.8 & 49.4 & 37.6 & 40.9 & 45.1 & 41.4 & 40.1 & 48.3 & 50.1 & 42.2 & 53.5 & 44.3 & 40.5 & 47.3 & 39.0 & 43.8\tabularnewline
Liu \etal \cite{DBLP:conf/eccv/LiuDZ0T20} ECCV'20 &  36.8 & 40.3 & 33.0 & 36.3 & 37.5 & 45.0 & 39.7 & 34.9 & 40.3 & 47.7 & 37.4 & 38.5 & 38.6 & 29.6 & 32.0 & 37.8\tabularnewline
Xu \etal \cite{DBLP:journals/corr/abs-2103-16385} CVPR'21 & 35.8 & 38.1 & 31.0 & 35.3 & 35.8 & 43.2 & 37.3 & 31.7 & 38.4 & 45.5 & 35.4 & 36.7 & 36.8 & \textbf{27.9} & 30.7 & 35.8\tabularnewline
\hline
Ours & \textbf{31.0} & \textbf{36.6} & 30.2 & \textbf{33.4} & 33.5 & 39.0 & \textbf{37.1} & \textbf{31.3} & \textbf{37.1} & 40.1 & \textbf{33.8} & \textbf{33.5} & \textbf{35.0} & 28.7 & 29.1 & \textbf{34.0}\tabularnewline
Ours + Refinement & 32.2 & 37.4 & \textbf{29.7} & 33.4 & \textbf{33.4} & \textbf{38.1} & 38.1 & 32.1 & 37.1 & 40.1 & 33.9 & 34.1 & 35.2 & 28.6 & \textbf{28.8} & 34.2\tabularnewline
\hline 
\end{tabular}

}
\end{center}\caption{Quantitative results on the Human3.6M \cite{DBLP:journals/pami/IonescuPOS14} dataset for single-frame 3D HPE with \textbf{(a)} CPN \cite{DBLP:conf/cvpr/ChenWPZYS18} predictions and \textbf{(b)} GT keypoints used as input. The evaluation is performed using MPJPE in $mm$ under \textit{Protocol \#1}. Methods denoted by (+) use extra data from MPII \cite{DBLP:conf/cvpr/AndrilukaPGS14}. Best results are shown in \textbf{bold}.}
\label{tab:results2}
\end{table*}

We compare the performance of our proposed model with recent state-of-the-art methods for single-frame 3D human pose estimation across two popular datasets: Human3.6M and MPI-INF-3DHP. We include both direct estimation and 2D-3D lifting methods in our analysis. Existing lifting methods use either multilayer perceptron or graph convolutional models to map from 2D to 3D. In contrast, the Joint Transformer uses a transformer architecture with intermediate supervision, error prediction and refinement. To highlight the significance of our improvements, we also compare to a "Base Transformer" architecture that consists of a linear embedding, spatial positional encoding \cite{zheng20213d} and transformer encoder blocks.

We first investigate the in-the-wild performance of our network where it takes the 2D predictions of a 2D pose detector as input. Table \ref{tab:results2}(a) shows the results on the Human3.6M dataset on 2D predictions of the cascaded pyramid network (CPN). Our Joint Transformer outperforms the state-of-the-art approaches by more than $1.4$ $mm$. The addition of the Refinement Transformer decreases our error a further $0.4$ $mm$ for a total improvement of $1.8$ $mm$. Our Joint Transformer also outperforms the Baseline Transformer without intermediate supervision, error prediction and with linear pose embedding by more than $3$ $mm$.

Since the above results are dependent on the quality of the output of the 2D pose detector, it is useful to remove the effect of the 2D backbone to characterise the theoretical performance of our network. Table \ref{tab:results2}(b) presents results on the Human3.6M dataset with ground truth 2D joint positions passed as input. Here, we find that our Joint Transformer surpasses the state of the art by $1.8$ $mm$. 
In this case, we find that our Refinement Transformer does not further decrease the prediction error. In fact, the performance of the Refinement Transformer seems correlated to the magnitude of the error prediction of the Joint Transformer. In the case of the ground-truth 2D inputs the predicted error is quite small, whilst the larger magnitude of the error prediction for the CPN keypoints can be effectively used by the Refinement Transformer to achieve a better prediction. In comparison to the state of the art, our method only uses $1.48$M parameters, compared to $3.70$M and $4.38$M parameters for the second- \cite{DBLP:journals/corr/abs-2103-16385} and third- \cite{DBLP:conf/eccv/LiuDZ0T20} placed methods respectively.

To show the generalisation ability of our method, we evaluate our Joint Transformer trained on Human3.6M directly on the MPI-INF-3DHP test dataset without any additional training. As shown in Table \ref{tab:results3}, we outperform the previous state of the art, even though some previous methods use additional training data from MPI-INF-3DHP. We also present the performance of our Joint Transformer followed by the Refinement Transformer. As can be seen, the Refinement Transformer further enhances our results by an additional small margin. This shows the ability of our method to generalise to unseen datasets. 
% \begin{itemize}
%     \item Show attention maps with intermediate and without if there are differences
% \end{itemize}
\begin{table*}[t]
\begin{center} 
\def\arraystretch{1.2} 
\resizebox{0.8\textwidth}{!}{

\begin{tabular}{ccccccc}
\hline 
 & Training data & GS (PCK) & noGS (PCK) & Outdoor (PCK) & All (PCK) & All (AUC)\tabularnewline
\hline 
\hline 
Martinez \etal \cite{martinez2017simple} ICCV'17 & H3.6M & 49.8 & 42.5 & 31.2 & 42.5 & 17.0 \tabularnewline
Mehta \etal \cite{8374605} 3DV'17 & H3.6M & 70.8 & 62.3 & 58.8 & 64.7 & 31.7 \tabularnewline
Yang \etal \cite{yang20183d} CVPR'18 & H3.6M+MPII & - & - & - & 69.0 & 32.0 \tabularnewline
Zhou \etal \cite{Zhou_2017_ICCV} ICCV'17 & H3.6M+MPII & 71.1 & 64.7 & 72.7 & 69.2 & 32.5 \tabularnewline
Luo \etal \cite{DBLP:conf/bmvc/LuoCY18} BMVC'18 & H3.6M & 71.3 & 59.4 & 65.7 & 65.6 & 33.2 \tabularnewline
Ci \etal \cite{ci2019optimizing} ICCV'19 & H3.6M & 74.8 & 70.8 & 77.3 & 74.0 & 36.7 \tabularnewline
Zhou \etal \cite{DBLP:conf/iccv/ZhouHJJL19} ICCV'19 & H3.6M+MPII & 75.6 & 71.3 & 80.3 & 75.3 & 38.0 \tabularnewline
Xu \etal \cite{DBLP:journals/corr/abs-2103-16385} CVPR'21 & H3.6M & 81.5 & 81.7 & 75.2 & 80.1 & 45.8 \tabularnewline
\hline
Base Transformer & H3.6M & 83.3 & 81.6 & 81.3 & 82.1 & 49.2\tabularnewline
Ours & H3.6M & 83.8 & 83.8 & 81.4 & 83.0 & 51.2 \tabularnewline
Ours + Refinement & H3.6M & \textbf{84.2} & \textbf{85.4} & \textbf{81.4} & \textbf{83.7} & \textbf{52.4} \tabularnewline
\hline 
\end{tabular}

}
\end{center}
\caption{Quantitative results for PCK and AUC on the MPI-INF-3DHP \cite{8374605} test dataset with GT keypoints used as input. Best results are shown in \textbf{bold}. The Jointformer is not trained on this dataset, showing the generalisation ability of the proposed method to unseen data.}
\label{tab:results3}
\end{table*}

\subsection{Ablation}
\label{sec:ablation}
% Preview source code for paragraph 5

\begin{table*}
\begin{center}
\def\arraystretch{1.2} 
\resizebox{0.8\textwidth}{!}{

\begin{tabular}{c|c|c||c|c|c||c|c||c||c||c|c||c||c}
\hline 
\multicolumn{3}{c||}{\textbf{$T_{layers}$}} & \multicolumn{3}{c||}{\textbf{$D_{h}$}} & \multicolumn{2}{c||}{\textbf{$Embed_{Pose}$}} & \multirow{2}{*}{\textbf{Int Sup.}} & \multirow{2}{*}{\textbf{Err Pred.}} & \multicolumn{2}{c||}{\textbf{Pos Enc.}} & \multicolumn{2}{c}{\textbf{Scores}}\tabularnewline
\cline{1-8} \cline{2-8} \cline{3-8} \cline{4-8} \cline{5-8} \cline{6-8} \cline{7-8} \cline{8-8} \cline{11-14} \cline{12-14} \cline{13-14} \cline{14-14} 
\textbf{4} & \textbf{5} & \textbf{6} & \textbf{64} & \textbf{128} & \textbf{256} & \textbf{Conv1D} & \textbf{Linear} &  &  & \textbf{Spatial} & \textbf{Frequency} & \textbf{MPJPE} & \textbf{P-MPJPE}\tabularnewline
\hline 
\hline 
\textbf{$\checkmark$} &  &  & \textbf{$\checkmark$} &  &  & \textbf{$\checkmark$} &  &  &  &  & & 44.2 & 33.7\tabularnewline
\hline 
\hline 
\textbf{$\checkmark$} &  &  & \textbf{$\checkmark$} &  &  & \textbf{$\checkmark$} &  & $\checkmark$ &  &  & & 34.5 & 26.5\tabularnewline
\hline 
\hline 
$\checkmark$ &  &  & $\checkmark$ &  &  & $\checkmark$ &  & $\checkmark$ & $\checkmark$ &  & & \textbf{34.0} & \textbf{26.5}\tabularnewline
\hline 
% \hline 
\textbf{$\checkmark$} &  &  & \textbf{$\checkmark$} &  &  & \textbf{$\checkmark$} &  & \textbf{$\checkmark$} & \textbf{$\checkmark$} & \textbf{$\checkmark$} & & 35.5 & 27.1\tabularnewline
\hline
\textbf{$\checkmark$} &  &  & \textbf{$\checkmark$} &  &  & \textbf{$\checkmark$} &  & \textbf{$\checkmark$} & \textbf{$\checkmark$} & & \textbf{$\checkmark$} & 35.2 & 27.0\tabularnewline
\hline 
\hline 
 & \textbf{$\checkmark$} &  & \textbf{$\checkmark$} &  &  & \textbf{$\checkmark$} &  & \textbf{$\checkmark$} & \textbf{$\checkmark$} &  & & 34.6 & 26.7\tabularnewline
\hline 
 &  & \textbf{$\checkmark$} & \textbf{$\checkmark$} &  &  & \textbf{$\checkmark$} &  & \textbf{$\checkmark$} & \textbf{$\checkmark$} &  & & 34.8 & 26.9\tabularnewline
\hline 
\hline 
\textbf{$\checkmark$} &  &  &  & \textbf{$\checkmark$} &  & \textbf{$\checkmark$} &  & \textbf{$\checkmark$} & \textbf{$\checkmark$} &  & & 34.4 & 26.9\tabularnewline
\hline 
\textbf{$\checkmark$} &  &  &  &  & \textbf{$\checkmark$} & \textbf{$\checkmark$} &  & \textbf{$\checkmark$} & \textbf{$\checkmark$} &  & & 34.3 & 26.5\tabularnewline
\hline 
\hline 
\textbf{$\checkmark$} &  &  & \textbf{$\checkmark$} &  &  &  & \textbf{$\checkmark$} & \textbf{$\checkmark$} & \textbf{$\checkmark$} &  & & 38.1 & 28.7\tabularnewline
\hline
\end{tabular}

}

\end{center}

\caption{Ablation study with quantitative results in $mm$ on the Human3.6M dataset \cite{DBLP:journals/pami/IonescuPOS14}. As input, GT keypoints are used. We start with
a basic Jointformer with \textbf{$T_{layers}=4$}, $D_{h}=64$ and \textbf{$Embed_{Pose}=Conv1D$
}(first row)\textbf{ }and next add intermediate supervision (second
row), error prediction (third row) and positional encoding (fourth and fifth row). We notice improvements with Int Sup. and Error Pred. but Pos
Enc. negatively impacts results. Next, with Int Sup. and Error Pred. added to the basic Jointformer, we experiment with
$T_{layers}$ (fifth and sixth rows), $D_{h}$ (seventh and eighth
rows) and \textbf{$Embed_{Pose}=Linear$} (ninth row). We obtain the
best results at $T_{layers}=4$, $D_{h}=64$ , \textbf{$Embed_{Pose}=Conv1D$,
}without position encoding and with Int Sup. and Err Pred.}
\label{tab:ablation}
\end{table*}

In this section, we evaluate the effectiveness of our modifications to the standard transformer architecture on the Human3.6M dataset \cite{DBLP:journals/pami/IonescuPOS14}. The results are reported in Table \ref{tab:ablation}. To remove the influence of the 2D pose detector, the results are reported using $17$ 2D ground truth joints as input. The purpose of the ablation study is to investigate the effects of the number of transformer layers ($T_{layers}$), size of the encoder ($D_h$), type of pose embedding used ($Embed_{Pose}$), intermediate supervision (Int Sup.), error prediction (Err Pred.) and positional encoding (Pos Enc.). All reported errors are the average of $3$ training runs. 
In the upper part of the table we observe the impact of the main contributions of this paper. Just by adding the intermediate supervision, we drastically improve the performance. Adding the error prediction and training the network in a multi-task framework further improves the results. We also observe that explicitly adding positional encoding to the transformer tokens reduces the performance. Since the order of the joints never changes, their positions are already implicitly encoded and further addition of an explicit encoding seems to confuse the network somewhat. This is the case for both the frequency based positional encoding, as well as the spatial encoding \cite{zheng20213d}.  
In the lower part of the table, we show the effect on a different number of transformer encoders and size of the hidden dimension. We show that the highest performance and most stable training can be achieved by using $4$ layers and a hidden dimension of $64$. While a larger number in either of these parameters increases the capacity of the network, the larger capacity seems not to be needed for best results. Finally, we can also observe that using a 1D convolution to embed the 2D inputs to the hidden dimension instead of a linear layer not only saves some network parameters, but also leads to better performance.
% \vspace{-0.2cm}
% We observe that using a 1D convolution with intermediate supervision and error prediction improves the performance significantly. On the other hand, positional encoding worsens the overall performance. Increasing the number of layers improves the performance slightly but we noticed the training to be more stable with 4 layers. 
% We start with a plain Jointformer network without error prediction and intermediate supervision. Firstly, we investigate the effect of intermediate supervision used between the stacked encoders. We observe a relative improvement in performance. The intermediate supervision forces each encoder to learn the full mapping. Secondly, we investigate the effect of error prediction. A further improvement is observed compared with the baseline. 

\section{Limitations}
\label{sec:limitations}

While the proposed system is robust to GT keypoints and robust detectors, we expect that the use of a noisy keypoint detector will contribute highly to poor performance. The Refinement Transformer improves results for the single-frame scenario. However, it is not trained on temporal noise and as such temporal inconsistencies will propagate to the estimated 3D poses.

\section{Conclusion}
\label{sec:conclusion}

In this work, we have presented a novel architecture for 3D pose estimation to analyse the previously-unexplored performance of transformers on single images. We demonstrate that implicitly encoding kinematic joint structure allows for a robust learning framework that outperforms the state of the art on established benchmarks. While the current approach addresses only single-frame input, it may be used as a modular component of temporal or multi-frame architectures in subsequent research.

% Despite outperforming some multi-frame state-of-the-art models, the current approach addresses only single-frame input. Future work,  As such, we present the given system as a modular component which may be naturally extended to temporal or multi-frame architectures in subsequent research.

% propose an amazing single frame pose estimator which outperforms state of the art. we do so by implementing x,y and z.
% As stated, we target a natural extension to temporal info, we expect that the presented work could be naturally extended to the temporal domain.  
% (extend to video, extend to absolute space)

\section{Acknowledgements}
This publication has emanated from research conducted with the financial support of Science Foundation Ireland (SFI) under the Grant Number 15/RP/2776.

% {\small
% \bibliography{egbib}}
\bibliographystyle{IEEEtran}
\bibliography{egbib}

\begin{thebibliography}{10}
\providecommand{\url}[1]{#1}
\csname url@samestyle\endcsname
\providecommand{\newblock}{\relax}
\providecommand{\bibinfo}[2]{#2}
\providecommand{\BIBentrySTDinterwordspacing}{\spaceskip=0pt\relax}
\providecommand{\BIBentryALTinterwordstretchfactor}{4}
\providecommand{\BIBentryALTinterwordspacing}{\spaceskip=\fontdimen2\font plus
\BIBentryALTinterwordstretchfactor\fontdimen3\font minus
  \fontdimen4\font\relax}
\providecommand{\BIBforeignlanguage}[2]{{%
\expandafter\ifx\csname l@#1\endcsname\relax
\typeout{** WARNING: IEEEtran.bst: No hyphenation pattern has been}%
\typeout{** loaded for the language `#1'. Using the pattern for}%
\typeout{** the default language instead.}%
\else
\language=\csname l@#1\endcsname
\fi
#2}}
\providecommand{\BIBdecl}{\relax}
\BIBdecl

\bibitem{davis1991gait}
R.~B. Davis~III, S.~Ounpuu, D.~Tyburski, and J.~R. Gage, ``A gait analysis data
  collection and reduction technique,'' \emph{Human movement science}, vol.~10,
  no.~5, pp. 575--587, 1991.

\bibitem{zheng20213d}
C.~Zheng, S.~Zhu, M.~Mendieta, T.~Yang, C.~Chen, and Z.~Ding, ``3d human pose
  estimation with spatial and temporal transformers,'' \emph{arXiv preprint
  arXiv:2103.10455}, 2021.

\bibitem{pavlakos2017coarse}
G.~Pavlakos, X.~Zhou, K.~G. Derpanis, and K.~Daniilidis, ``Coarse-to-fine
  volumetric prediction for single-image 3d human pose,'' in \emph{Proceedings
  of the IEEE Conference on Computer Vision and Pattern Recognition}, 2017, pp.
  7025--7034.

\bibitem{cao2019openpose}
Z.~Cao, G.~Hidalgo, T.~Simon, S.-E. Wei, and Y.~Sheikh, ``Openpose: realtime
  multi-person 2d pose estimation using part affinity fields,'' \emph{IEEE
  transactions on pattern analysis and machine intelligence}, vol.~43, no.~1,
  pp. 172--186, 2019.

\bibitem{sun2019deep}
K.~Sun, B.~Xiao, D.~Liu, and J.~Wang, ``Deep high-resolution representation
  learning for human pose estimation,'' in \emph{Proceedings of the IEEE/CVF
  Conference on Computer Vision and Pattern Recognition}, 2019, pp. 5693--5703.

\bibitem{martinez2017simple}
J.~Martinez, R.~Hossain, J.~Romero, and J.~J. Little, ``A simple yet effective
  baseline for 3d human pose estimation,'' in \emph{Proceedings of the IEEE
  International Conference on Computer Vision}, 2017, pp. 2640--2649.

\bibitem{zhao2019semantic}
L.~Zhao, X.~Peng, Y.~Tian, M.~Kapadia, and D.~N. Metaxas, ``Semantic graph
  convolutional networks for 3d human pose regression,'' in \emph{Proceedings
  of the IEEE/CVF Conference on Computer Vision and Pattern Recognition}, 2019,
  pp. 3425--3435.

\bibitem{DBLP:conf/eccv/LiuDZ0T20}
K.~Liu, R.~Ding, Z.~Zou, L.~Wang, and W.~Tang, ``A comprehensive study of
  weight sharing in graph networks for 3d human pose estimation,'' in
  \emph{Computer Vision - {ECCV} 2020 - 16th European Conference, Glasgow, UK,
  August 23-28, 2020, Proceedings, Part {X}}, ser. Lecture Notes in Computer
  Science, A.~Vedaldi, H.~Bischof, T.~Brox, and J.~Frahm, Eds., vol.
  12355.\hskip 1em plus 0.5em minus 0.4em\relax Springer, 2020, pp. 318--334.

\bibitem{zhou2020towards}
K.~Zhou, X.~Huang, Y.~Li, D.~Zha, R.~Chen, and X.~Hu, ``Towards deeper graph
  neural networks with differentiable group normalization,'' \emph{arXiv
  preprint arXiv:2006.06972}, 2020.

\bibitem{vaswani2017attention}
A.~Vaswani, N.~Shazeer, N.~Parmar, J.~Uszkoreit, L.~Jones, A.~N. Gomez,
  L.~Kaiser, and I.~Polosukhin, ``Attention is all you need,'' \emph{arXiv
  preprint arXiv:1706.03762}, 2017.

\bibitem{dosovitskiy2020image}
A.~Dosovitskiy, L.~Beyer, A.~Kolesnikov, D.~Weissenborn, X.~Zhai,
  T.~Unterthiner, M.~Dehghani, M.~Minderer, G.~Heigold, S.~Gelly \emph{et~al.},
  ``An image is worth 16x16 words: Transformers for image recognition at
  scale,'' \emph{arXiv preprint arXiv:2010.11929}, 2020.

\bibitem{carion2020end}
N.~Carion, F.~Massa, G.~Synnaeve, N.~Usunier, A.~Kirillov, and S.~Zagoruyko,
  ``End-to-end object detection with transformers,'' in \emph{European
  Conference on Computer Vision}.\hskip 1em plus 0.5em minus 0.4em\relax
  Springer, 2020, pp. 213--229.

\bibitem{sharir2021image}
G.~Sharir, A.~Noy, and L.~Zelnik-Manor, ``An image is worth 16x16 words, what
  is a video worth?'' \emph{arXiv preprint arXiv:2103.13915}, 2021.

\bibitem{lin2020end}
K.~Lin, L.~Wang, and Z.~Liu, ``End-to-end human pose and mesh reconstruction
  with transformers,'' \emph{arXiv preprint arXiv:2012.09760}, 2020.

\bibitem{li2021lifting}
W.~Li, H.~Liu, R.~Ding, M.~Liu, and P.~Wang, ``Lifting transformer for 3d human
  pose estimation in video,'' \emph{arXiv preprint arXiv:2103.14304}, 2021.

\bibitem{DBLP:conf/cvpr/PavlakosZD18}
G.~Pavlakos, X.~Zhou, and K.~Daniilidis, ``Ordinal depth supervision for 3d
  human pose estimation,'' in \emph{2018 {IEEE} Conference on Computer Vision
  and Pattern Recognition, {CVPR} 2018, Salt Lake City, UT, USA, June 18-22,
  2018}.\hskip 1em plus 0.5em minus 0.4em\relax {IEEE} Computer Society, 2018,
  pp. 7307--7316.

\bibitem{DBLP:conf/iccv/0001SLW17}
X.~Sun, J.~Shang, S.~Liang, and Y.~Wei, ``Compositional human pose
  regression,'' in \emph{{IEEE} International Conference on Computer Vision,
  {ICCV} 2017, Venice, Italy, October 22-29, 2017}.\hskip 1em plus 0.5em minus
  0.4em\relax {IEEE} Computer Society, 2017, pp. 2621--2630.

\bibitem{ci2019optimizing}
H.~Ci, C.~Wang, X.~Ma, and Y.~Wang, ``Optimizing network structure for 3d human
  pose estimation,'' in \emph{Proceedings of the IEEE/CVF International
  Conference on Computer Vision}, 2019, pp. 2262--2271.

\bibitem{DBLP:journals/corr/abs-2103-16385}
T.~Xu and W.~Takano, ``Graph stacked hourglass networks for 3d human pose
  estimation,'' \emph{CoRR}, vol. abs/2103.16385, 2021.

\bibitem{DBLP:conf/iccv/TekinMSF17}
B.~Tekin, P.~M{\'{a}}rquez{-}Neila, M.~Salzmann, and P.~Fua, ``Learning to fuse
  2d and 3d image cues for monocular body pose estimation,'' in \emph{{IEEE}
  International Conference on Computer Vision, {ICCV} 2017, Venice, Italy,
  October 22-29, 2017}.\hskip 1em plus 0.5em minus 0.4em\relax {IEEE} Computer
  Society, 2017, pp. 3961--3970.

\bibitem{yang20183d}
W.~Yang, W.~Ouyang, X.~Wang, J.~Ren, H.~Li, and X.~Wang, ``3d human pose
  estimation in the wild by adversarial learning,'' in \emph{Proceedings of the
  IEEE Conference on Computer Vision and Pattern Recognition}, 2018, pp.
  5255--5264.

\bibitem{DBLP:conf/aaai/FangXWLZ18}
H.~Fang, Y.~Xu, W.~Wang, X.~Liu, and S.~Zhu, ``Learning pose grammar to encode
  human body configuration for 3d pose estimation,'' in \emph{Proceedings of
  the Thirty-Second {AAAI} Conference on Artificial Intelligence, (AAAI-18),
  the 30th innovative Applications of Artificial Intelligence (IAAI-18), and
  the 8th {AAAI} Symposium on Educational Advances in Artificial Intelligence
  (EAAI-18), New Orleans, Louisiana, USA, February 2-7, 2018}, S.~A. McIlraith
  and K.~Q. Weinberger, Eds.\hskip 1em plus 0.5em minus 0.4em\relax {AAAI}
  Press, 2018, pp. 6821--6828.

\bibitem{DBLP:conf/iccv/SharmaVBSJ19}
S.~Sharma, P.~T. Varigonda, P.~Bindal, A.~Sharma, and A.~Jain, ``Monocular 3d
  human pose estimation by generation and ordinal ranking,'' in \emph{2019
  {IEEE/CVF} International Conference on Computer Vision, {ICCV} 2019, Seoul,
  Korea (South), October 27 - November 2, 2019}.\hskip 1em plus 0.5em minus
  0.4em\relax {IEEE}, 2019, pp. 2325--2334.

\bibitem{DBLP:conf/iccv/ZhouHJJL19}
K.~Zhou, X.~Han, N.~Jiang, K.~Jia, and J.~Lu, ``Hemlets pose: Learning
  part-centric heatmap triplets for accurate 3d human pose estimation,'' in
  \emph{2019 {IEEE/CVF} International Conference on Computer Vision, {ICCV}
  2019, Seoul, Korea (South), October 27 - November 2, 2019}.\hskip 1em plus
  0.5em minus 0.4em\relax {IEEE}, 2019, pp. 2344--2353.

\bibitem{lecun1989backpropagation}
Y.~LeCun, B.~Boser, J.~S. Denker, D.~Henderson, R.~E. Howard, W.~Hubbard, and
  L.~D. Jackel, ``Backpropagation applied to handwritten zip code
  recognition,'' \emph{Neural computation}, vol.~1, no.~4, pp. 541--551, 1989.

\bibitem{wang2018non}
X.~Wang, R.~Girshick, A.~Gupta, and K.~He, ``Non-local neural networks,'' in
  \emph{Proceedings of the IEEE conference on computer vision and pattern
  recognition}, 2018, pp. 7794--7803.

\bibitem{ramachandran2019stand}
P.~Ramachandran, N.~Parmar, A.~Vaswani, I.~Bello, A.~Levskaya, and J.~Shlens,
  ``Stand-alone self-attention in vision models,'' \emph{arXiv preprint
  arXiv:1906.05909}, 2019.

\bibitem{zhao2020exploring}
H.~Zhao, J.~Jia, and V.~Koltun, ``Exploring self-attention for image
  recognition,'' in \emph{Proceedings of the IEEE/CVF Conference on Computer
  Vision and Pattern Recognition}, 2020, pp. 10\,076--10\,085.

\bibitem{arnab2021vivit}
A.~Arnab, M.~Dehghani, G.~Heigold, C.~Sun, M.~Lu{\v{c}}i{\'c}, and C.~Schmid,
  ``Vivit: A video vision transformer,'' \emph{arXiv preprint
  arXiv:2103.15691}, 2021.

\bibitem{ruggero2017benchmarking}
M.~Ruggero~Ronchi and P.~Perona, ``Benchmarking and error diagnosis in
  multi-instance pose estimation,'' in \emph{Proceedings of the IEEE
  international conference on computer vision}, 2017, pp. 369--378.

\bibitem{lin2014microsoft}
T.-Y. Lin, M.~Maire, S.~Belongie, J.~Hays, P.~Perona, D.~Ramanan,
  P.~Doll{\'a}r, and C.~L. Zitnick, ``Microsoft {COCO}: Common objects in
  context,'' in \emph{European conference on computer vision}.\hskip 1em plus
  0.5em minus 0.4em\relax Springer, 2014, pp. 740--755.

\bibitem{moon2019posefix}
G.~Moon, J.~Y. Chang, and K.~M. Lee, ``Posefix: Model-agnostic general human
  pose refinement network,'' in \emph{Proceedings of the IEEE/CVF Conference on
  Computer Vision and Pattern Recognition}, 2019, pp. 7773--7781.

\bibitem{chang2019poselifter}
J.~Y. Chang, G.~Moon, and K.~M. Lee, ``Poselifter: Absolute 3d human pose
  lifting network from a single noisy 2d human pose,'' \emph{arXiv preprint
  arXiv:1910.12029}, 2019.

\bibitem{fieraru2018learning}
M.~Fieraru, A.~Khoreva, L.~Pishchulin, and B.~Schiele, ``Learning to refine
  human pose estimation,'' in \emph{Proceedings of the IEEE conference on
  computer vision and pattern recognition workshops}, 2018, pp. 205--214.

\bibitem{zhang2019human}
H.~Zhang, H.~Ouyang, S.~Liu, X.~Qi, X.~Shen, R.~Yang, and J.~Jia, ``Human pose
  estimation with spatial contextual information,'' \emph{arXiv preprint
  arXiv:1901.01760}, 2019.

\bibitem{wang2020graph}
J.~Wang, X.~Long, Y.~Gao, E.~Ding, and S.~Wen, ``Graph-{PCNN}: Two stage human
  pose estimation with graph pose refinement,'' in \emph{European Conference on
  Computer Vision}.\hskip 1em plus 0.5em minus 0.4em\relax Springer, 2020, pp.
  492--508.

\bibitem{newell2016stacked}
A.~Newell, K.~Yang, and J.~Deng, ``Stacked hourglass networks for human pose
  estimation,'' in \emph{European conference on computer vision}.\hskip 1em
  plus 0.5em minus 0.4em\relax Springer, 2016, pp. 483--499.

\bibitem{DBLP:conf/nips/PaszkeGMLBCKLGA19}
A.~Paszke, S.~Gross, F.~Massa, A.~Lerer, J.~Bradbury, G.~Chanan, T.~Killeen,
  Z.~Lin, N.~Gimelshein, L.~Antiga, A.~Desmaison, A.~K{\"{o}}pf, E.~Yang,
  Z.~DeVito, M.~Raison, A.~Tejani, S.~Chilamkurthy, B.~Steiner, L.~Fang,
  J.~Bai, and S.~Chintala, ``Pytorch: An imperative style, high-performance
  deep learning library,'' in \emph{Advances in Neural Information Processing
  Systems 32: Annual Conference on Neural Information Processing Systems 2019,
  NeurIPS 2019, December 8-14, 2019, Vancouver, BC, Canada}, H.~M. Wallach,
  H.~Larochelle, A.~Beygelzimer, F.~d'Alch{\'{e}}{-}Buc, E.~B. Fox, and
  R.~Garnett, Eds., 2019, pp. 8024--8035.

\bibitem{DBLP:conf/iclr/LoshchilovH19}
I.~Loshchilov and F.~Hutter, ``Decoupled weight decay regularization,'' in
  \emph{7th International Conference on Learning Representations, {ICLR} 2019,
  New Orleans, LA, USA, May 6-9, 2019}.\hskip 1em plus 0.5em minus 0.4em\relax
  OpenReview.net, 2019.

\bibitem{DBLP:conf/iclr/LoshchilovH17}
------, ``{SGDR:} stochastic gradient descent with warm restarts,'' in
  \emph{5th International Conference on Learning Representations, {ICLR} 2017,
  Toulon, France, April 24-26, 2017, Conference Track Proceedings}.\hskip 1em
  plus 0.5em minus 0.4em\relax OpenReview.net, 2017.

\bibitem{pavllo20193d}
D.~Pavllo, C.~Feichtenhofer, D.~Grangier, and M.~Auli, ``3d human pose
  estimation in video with temporal convolutions and semi-supervised
  training,'' in \emph{Proceedings of the IEEE/CVF Conference on Computer
  Vision and Pattern Recognition}, 2019, pp. 7753--7762.

\bibitem{DBLP:journals/pami/IonescuPOS14}
C.~Ionescu, D.~Papava, V.~Olaru, and C.~Sminchisescu, ``Human3.6m: Large scale
  datasets and predictive methods for 3d human sensing in natural
  environments,'' \emph{{IEEE} Trans. Pattern Anal. Mach. Intell.}, vol.~36,
  no.~7, pp. 1325--1339, 2014.

\bibitem{8374605}
D.~Mehta, H.~Rhodin, D.~Casas, P.~Fua, O.~Sotnychenko, W.~Xu, and C.~Theobalt,
  ``Monocular 3d human pose estimation in the wild using improved cnn
  supervision,'' in \emph{2017 International Conference on 3D Vision (3DV)},
  2017, pp. 506--516.

\bibitem{moreno20173d}
F.~Moreno-Noguer, ``3d human pose estimation from a single image via distance
  matrix regression,'' in \emph{Proceedings of the IEEE Conference on Computer
  Vision and Pattern Recognition}, 2017, pp. 2823--2832.

\bibitem{DBLP:conf/bmvc/LuoCY18}
C.~Luo, X.~Chu, and A.~L. Yuille, ``Orinet: {A} fully convolutional network for
  3d human pose estimation,'' in \emph{British Machine Vision Conference 2018,
  {BMVC} 2018, Newcastle, UK, September 3-6, 2018}.\hskip 1em plus 0.5em minus
  0.4em\relax {BMVA} Press, 2018, p.~92.

\bibitem{DBLP:conf/cvpr/ChenWPZYS18}
Y.~Chen, Z.~Wang, Y.~Peng, Z.~Zhang, G.~Yu, and J.~Sun, ``Cascaded pyramid
  network for multi-person pose estimation,'' in \emph{2018 {IEEE} Conference
  on Computer Vision and Pattern Recognition, {CVPR} 2018, Salt Lake City, UT,
  USA, June 18-22, 2018}.\hskip 1em plus 0.5em minus 0.4em\relax {IEEE}
  Computer Society, 2018, pp. 7103--7112.

\bibitem{DBLP:conf/cvpr/AndrilukaPGS14}
M.~Andriluka, L.~Pishchulin, P.~V. Gehler, and B.~Schiele, ``2d human pose
  estimation: New benchmark and state of the art analysis,'' in \emph{2014
  {IEEE} Conference on Computer Vision and Pattern Recognition, {CVPR} 2014,
  Columbus, OH, USA, June 23-28, 2014}.\hskip 1em plus 0.5em minus 0.4em\relax
  {IEEE} Computer Society, 2014, pp. 3686--3693.

\bibitem{Zhou_2017_ICCV}
X.~Zhou, Q.~Huang, X.~Sun, X.~Xue, and Y.~Wei, ``Towards 3d human pose
  estimation in the wild: A weakly-supervised approach,'' in \emph{Proceedings
  of the IEEE International Conference on Computer Vision (ICCV)}, Oct 2017.

\end{thebibliography}

\end{document}